\documentclass[a4paper, 10 pt, conference]{ieeeconf}  

\IEEEoverridecommandlockouts                              

\usepackage{graphicx}
\graphicspath{ {Images/} }
\usepackage{hyperref}  

\title{\LARGE \bf
Human-centered manipulation and navigation \\ with Robot DE NIRO  
}

\author{Fabian Falck, Sagar Doshi, Nico Smuts, John Lingi, Kim Rants, Petar Kormushev
\thanks{The authors are affiliated with Imperial College London, Robot Intelligence Lab, Department of Computing and Dyson School of Design Engineering. {\tt\small fabian.falck17, sagar.doshi17, nicolaas.smuts17, john.lingi09, kim.rants17, p.kormushev@imperial.ac.uk} }
}

\begin{document}

\maketitle
\thispagestyle{empty}
\pagestyle{empty}

\begin{abstract}

Social assistance robots in health and elderly care have the potential to support and ease human lives. Given the macrosocial trends of aging and long-lived populations, robotics-based care research mainly focused on helping the elderly live independently. In this paper, we introduce Robot DE NIRO, a research platform that aims to support the supporter (the caregiver) and also offers direct human-robot interaction for the care recipient. Augmented by several sensors, DE NIRO is capable of complex manipulation tasks. It reliably interacts with humans and can autonomously and swiftly navigate through dynamically changing environments. We describe preliminary experiments in a demonstrative scenario and discuss DE NIRO's design and capabilities. We put particular emphases on safe, human-centered interaction procedures implemented in both hardware and software, including collision avoidance in manipulation and navigation as well as an intuitive perception stack through speech and face recognition.

\end{abstract}

\section{Introduction}

Social assistance robots for elderly care or general nursing have been subject to extensive research in recent years. They may serve to counterbalance the global nursing shortage caused by both demand factors, such as demographic trends \cite{tapus2007grandChallenge}, and supply factors, such as unfavorable working environments or egregious wage disparities \cite{super2002will} \cite{oulton2006global}. Most systems are focused on directly assisting the care recipient -- often an independently living elderly person -- with social companionship or simple household services \cite{schroeter2013realization} \cite{fischinger2016hobbit}. However, elderly care today is still predominantly administered by human caregivers, who may themselves benefit from a robot assistant. Instead of seeking to replace caregivers, we propose Robot DE NIRO (Design Engineering's Natural Interaction RObot) \footnote{ Further information on DE NIRO can be found under \url{http://www.imperial.ac.uk/robot-intelligence/robots/robot_de_niro/}} as a tool for caregivers. DE NIRO is a collaborative research platform that can aid geriatric nurses by performing well-defined, repeated auxiliary tasks, such as retrieving a bottle of medicine and taking it to the care recipient. In designing DE NIRO, we have put an emphasis on natural and safe human-robot interaction procedures across multiple components, including speech and face recognition and collision avoidance.

This paper explains more of the context and functionalities of DE NIRO. In section II, we briefly discuss related work on care and social assistance robots. Section III gives an overview of DE NIRO's initial hardware design and the applied software frameworks. Section IV relates our preliminary experiments with DE NIRO. We give an overview of its current perception capabilities and possible actions, with focus on manipulation and LIDAR-based navigation. Finally, section V concludes the paper and gives an outlook for future work on this research platform.

\begin{figure}
\centering
\includegraphics[width=8cm]{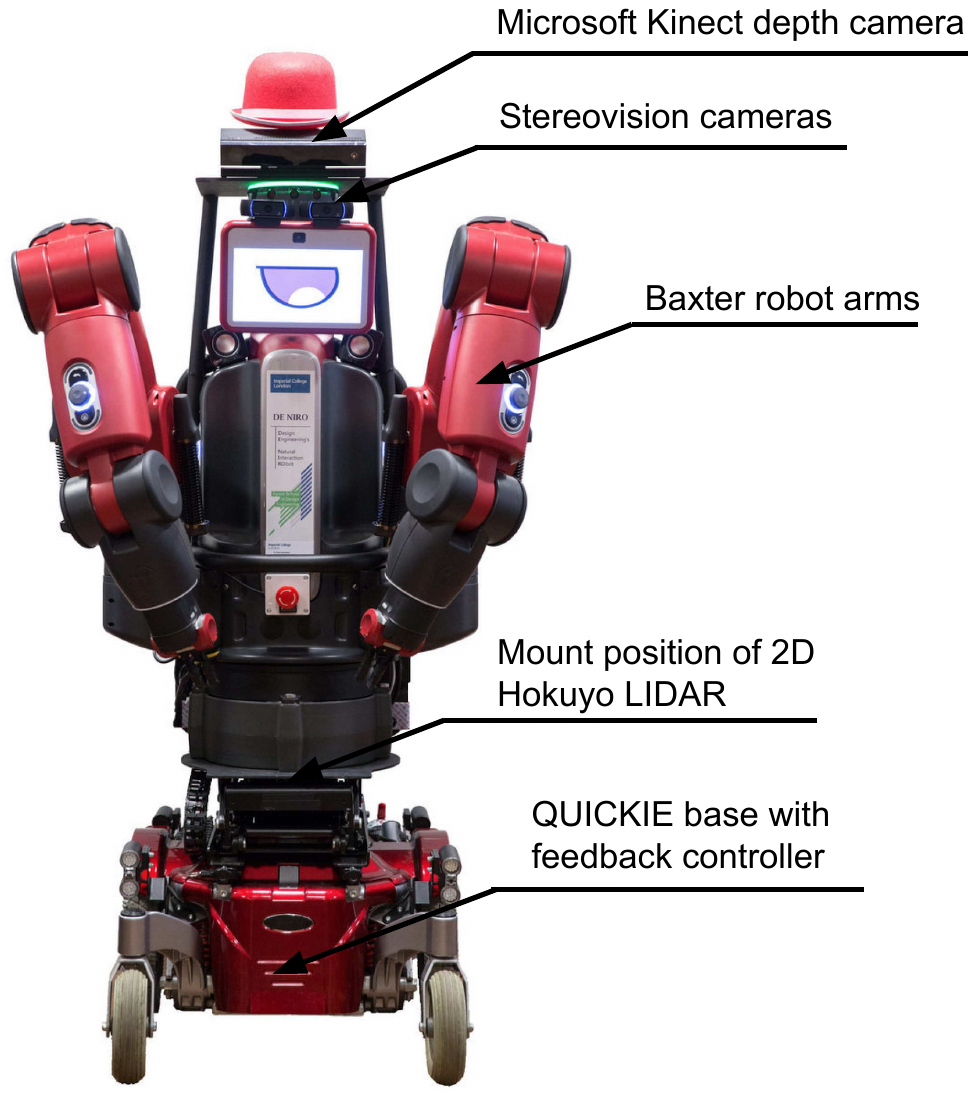}
\caption{Robot DE NIRO - a collaborative research platform for mobile manipulation. The figure shows its design, main components and sensors.} %
\label{fig:DENIRO}
\end{figure}

\section{Related Work}

In current literature, robot systems for elderly care are typically built as autonomous systems that directly support personal independence through social companionship, routine household services, and telepresence \cite{fischinger2016hobbit}. Examples of state-of-the-art platforms in service robotics -- some of them combining all three categories -- are Care-O-Bot 3 \cite{graf2009robotic}, ASIMO \cite{sakagami2002intelligent}, HRP-3 \cite{kaneko2008humanoid}, various solutions for ambient assisted living such as the DOMEO RobuMate \cite{domeo} and a social assistant robot for people with mild cognitive impairments \cite{schroeter2013realization} \cite{rashidi2013survey}. Willow Garage built the more general platform PR2, which has been used by various universities to build human-centered applications that include supporting the elderly \cite{cousins2010ros}. Advances for better social interaction and improved navigation have been made, such as gesture recognition for service ordering \cite{zhao2014kinect} or efficient navigation in unstructured household environments in an emergency situation \cite{berns2010use}. Target care recipients are those who suffer from psychological diseases or, more commonly, those with limited mobility, including people who are elderly, disabled, temporarily or chronically sick, pregnant, or otherwise constrained \cite{graf2004care}. 

With these recent advances in social assistant robots, public and academic debate has not yet settled on whether and for which purposes such systems are an ethical and desirable outcome for our society \cite{sparrow2006hands} \cite{wallach2008moral}. For instance, Sharkey and Sharkey point out six main ethical concerns with robot assistance for elderly care, including human isolation, loss of control and personal liberty, and deception and infantilization \cite{Sharkey2012}. Furthermore, \cite{zsiga2013home} note varying attitudes and preferences regarding social assistance robots. While this discussion is not settled, we believe it is approriate in the interim to focus robot assistance on the caregiver. This allows the caregiver to delegate simple, repetitive tasks to a social robot assistant and gain time for more complex, empathetic tasks.

Much less work has sought to support caregivers as they shoulder typical nursing tasks to care for elderly persons. Among those, \cite{Ding2017} propose a transfer assistant robot to lift a patient from a bed to a wheelchair through a model-based holding posture estimation and a model-free generation of the lifting motion. \cite{srinivasa2010herb} demonstrate complex manipulation skills for bottles and cans based on sparse 3D models for object recognition and pose estimation. They integrate these skills with navigation and mapping capabilities, both in a static and a dynamic environment. 
\section{Design}

DE NIRO's core design idea is to combine the industrial Baxter dual robot arms with autonomous navigation into a mobile manipulation research platform. The Baxter arms are a common standard in human-robot interaction and allow complex manipulation of objects \cite{baxter_arms}. A particular safety feature of the arms is their passive compliance through series elastic actuators. This allows the robot to interact with humans in close proximity to the robot safely, since in the case of a contact, most of the physical impact is absorbed. The Baxter arms are mounted on a QUICKIE movable electric wheelchair base. Its differential drive is operated with a custom PID angular position and velocity controller, allowing primitive motion commands for navigation. The controller itself is implemented through an integrated Mbed microcontroller \cite{quickie}. On the hardware side, multiple layers of safety are operating for the event of an emergency. An automated interrupt procedure stops the movement of QUICKIE if a time out appears. Furthermore, both on-board and wireless e-stop buttons allow the user to brake the robot immediately.

These core elements are augmented with the following sensors: a Microsoft Kinect RGB-D camera, various stereovision cameras with built-in microphones, ultrasonic and infrared proximity sensors, speakers for audio output, and a Hokuyo 2D LIDAR scanner. Equipped with this extensive list of sensors and actuators, DE NIRO is capable of performing a wide variety of the typical, repetitive tasks of a caregiver, such as serving drinks and food, grasping, fetching and carrying of objects, and helping others come to a standing position \cite{graf2004care}. Figure \ref{fig:DENIRO} visualizes the hardware design of DE NIRO as a whole.

To handle concurrent execution and both synchronous and asynchronous communication between components, we use Robot Operating System (ROS) as middleware. We define distinct functionalities of the robot with a finite-state machine, such as \verb|listening| (for command input) or \verb|grasping| (to physically pick up an object). The state machine handles the control flow among these states. Furthermore, we set up a wireless LAN network for concurrent communication between the Baxter core and the two controlling laptops mounted on the back of DE NIRO. For testing and debugging purposes and in order to integrate all sensor outputs and log messages, we built an rqt-based GUI illustrated in figure \ref{fig:markers_gui} and useful mainly to the technical user \cite{rqt}.

\section{Implementation and Experiments} 

Our primary work focused on the development and integration of state-of-the-art algorithms to perform a particular demonstration scenario. This scenario involved interacting with a user to receive a command indicating which object to grasp; navigating to and from an object warehouse; and grasping, manipulating, and passing back the requested object to the requester. 

\begin{figure}[t]
        \centering
        \includegraphics[width=0.5\textwidth]{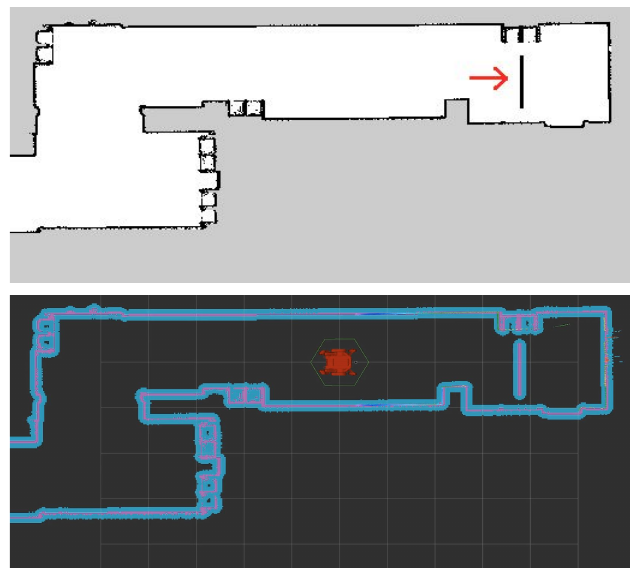} 
        \caption{A static map of a corridor (top). The red arrow points at a synthetic barrier manually added to the map. A costmap of 10cm surrounds all static barriers. The hexagonal shape is a synthetic barrier around the QUICKIE base used for collision avoidance (bottom).}
		\label{fig:static_maps}
\end{figure}

\textbf{Perception and user interaction.} One challenge was to recognize the user's face and distinguish that individual from others. To solve this, we used a pre-trained machine learning model based on the Residual Learning for Image Recognition (ResNet) approach that we applied to video frames retrieved by the Kinect camera \cite{resnet_paper} \cite{face_recog_interface}. The model has reached a 99.38\% accuracy on a standard benchmark \cite{dlib_resnet}. It compares the output vector encodings of known faces (via saved images) with others extracted from the processed frames by computing a distance metric between the saved and incoming vector encodings. If that distance is below a threshold, it predicts a positive match. We tuned the model to predict with a very low false-positive rate at the cost of a slightly increased false-negative rate, in order to be less vulnerable to unintended interactions.

To naturally interact with a user, we implemented a speech recognition system using the offline library \textit{CMU Sphinx} \cite{sphinx}. We defined a JSpeech Grammar to allow voice commands in a specific, yet flexible format tested on a variety of accents. Furthermore, the system regularly calibrates to background noise levels. Compared to several online APIs, this implementation achieved the most robust results in varying environments. For audio output, we elected to use \textit{eSpeak}, a simple speech package, over more sophisticated candidates, due to its high reliability, rapid response time, low required processing power, and -- especially -- offline implementation \cite{espeak}.

\begin{figure}[t!]  
        \centering
        \includegraphics[width=0.5\textwidth]{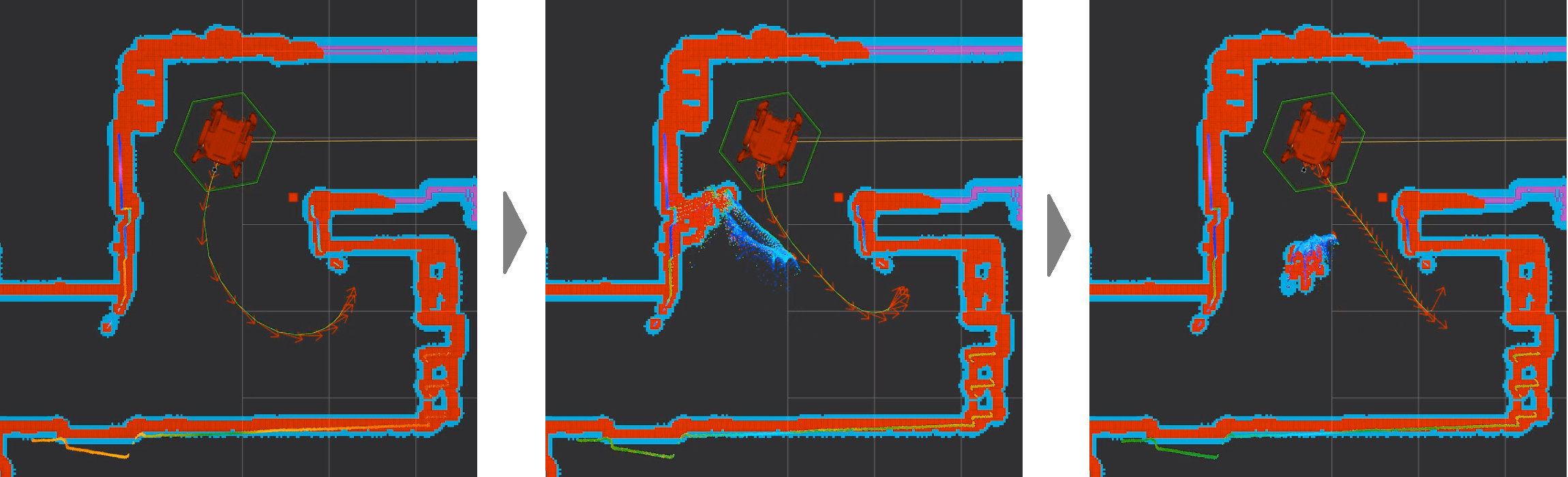}
        \caption{The original optimal path (left), encountering a dynamic obstacle (middle), and adjusting in response (right) during trajectory planning.}
		\label{fig:route_sequence}
\end{figure}

\begin{figure}[b!]
        \centering
        \includegraphics[width=0.5\textwidth]{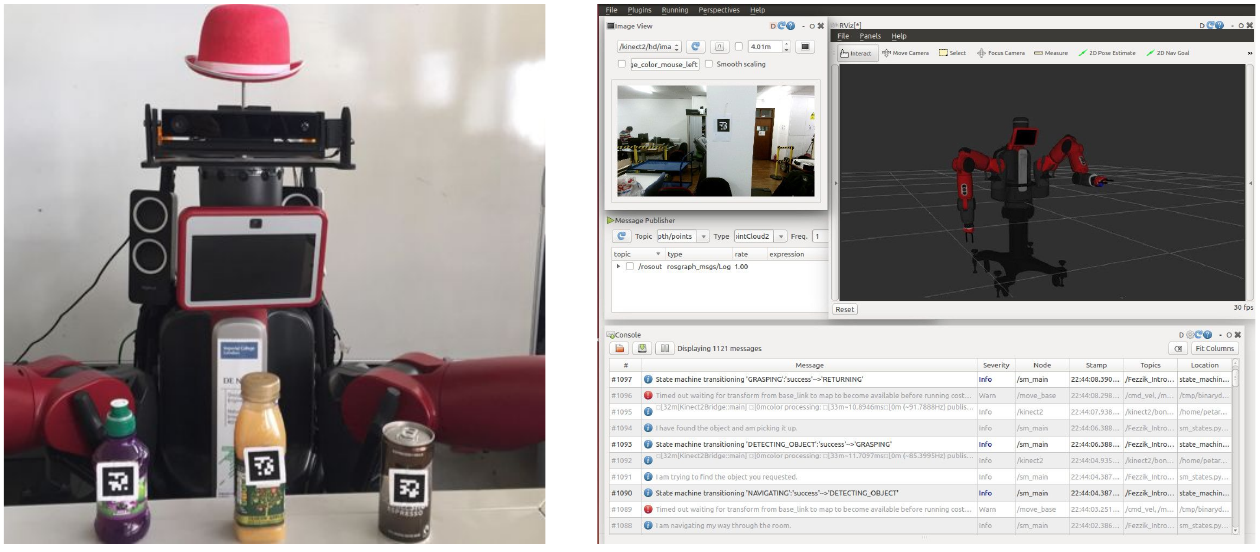} 
        \caption{The 2D fiducial markers attached to test objects (left) and the rqt-based technical GUI for testing purposes (right).}
		\label{fig:markers_gui}
\end{figure}

\textbf{Navigation, Mapping, and Planning.} We implemented a reliable navigation stack consisting of mapping, localization, and trajectory planning. To generate an initial static bitmap, we apply a SLAM-based approach to the LIDAR sensor in order to detect any spatial boundaries and 2D artifacts in a predefined space \cite{hectormapping}. Then, we localize the robot by overlaying a dynamic map onto the static map. This dynamic mapping feature is also useful for collision avoidance, which is particularly important when DE NIRO is in the vicinity of untrained humans. For additional safety, we impose a costmap to create a virtual cushion around all static and dynamic obstacles and around the QUICKIE base itself. The static map is illustrated in figure \ref{fig:static_maps}.

For efficient trajectory planning, we use a ``timed elastic band'' approach, conceiving of trajectory planning as a multi-objective optimization problem \cite{timedelasticband} \cite{teblocalplanner}. This approach minimizes the costs that are assigned to variables like total travel time and obstacle proximity simultaneously. This helps DE NIRO to maintain a safe distance from users. The planned linear and angular velocities of the optimal path are then scaled and smoothed by the custom PID controller discussed above. Finally, an electric signal to the motor produces actuating rotational movement \cite{quickie}. Figure \ref{fig:route_sequence} illustrates a path replanning scenario for when DE NIRO detects a dynamic obstacle.

\begin{figure}[t!]
        \centering
        \includegraphics[width=0.5\textwidth]{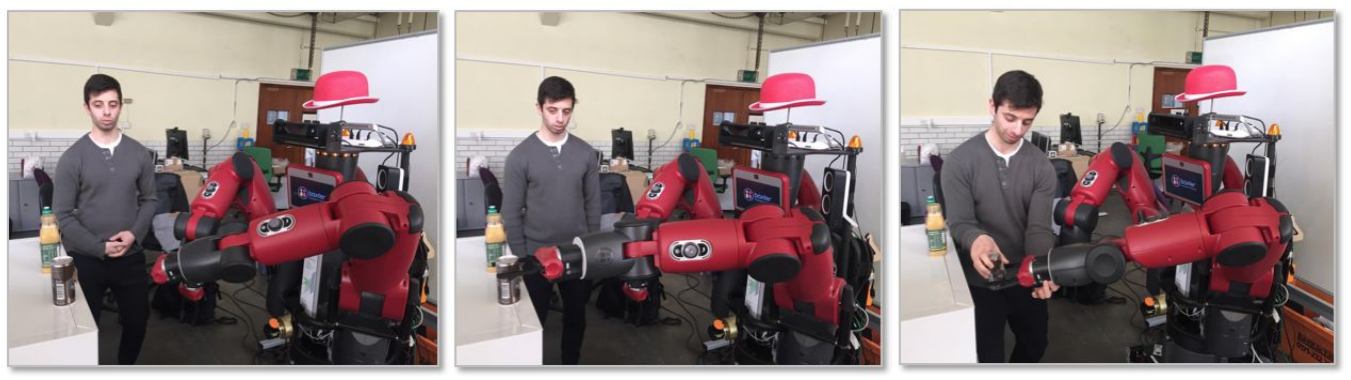} 
        \caption{The three grasping stages of moving to an intermediate position (left), grasping the object (middle), and executing handover-mode (right).}
		\label{fig:grasping}
\end{figure}

\textbf{Object Recognition and Manipulation.} To recognize the target object and localize it in 3D space, we relied on 2D fiducial markers attached to the object \cite{chili_tag}. We also experimented with various other more generic and scalable solutions, including recognizing objects on a planar surface. This solution and other attempts we made did not work robustly enough during grasping. The 2D fiducial markers depicted in figure \ref{fig:markers_gui} were much more consistent. 

To control the Baxter arms, we employed an inverse kinematics solver to compute each of the seven joint angle trajectories needed to reach an object \cite{ik_solver}. We designed a dynamic awareness procedure so that DE NIRO selects the most appropriate arm to make a grasp attempt; reacts to changes in the object's location during grasping; and actively avoids collisions, say, with the unused arm. We experimented with various constraints on possible joint angles and settled on a grasping procedure with one intermediate point that achieves its goal most frequently. After grasping the object, the user can retrieve it easily during \verb|handover mode| by imposing a small force along the z-axis. The experimental grasping process is illustrated in figure \ref{fig:grasping}.

\section{Conclusion}

In this research \footnote{ We open-sourced our object-oriented \textit{code} base in Python together with an extensive \textit{documentation} for it that will be continuously updated. Furthermore, we have published a \textit{video} illustrating some of the current core skills of DE NIRO. Publications of \textit{sensor data} from DE NIRO will follow soon. All resources are linked at \url{http://www.imperial.ac.uk/robot-intelligence/software/}.}, we presented the design and implementation of Robot DE NIRO to support geriatric nurses in interaction tasks with care recipients. DE NIRO's (current) design is limited in various ways: First, the robot design is nonholonomic, being limited to only forward and backward translational and rotational (but no side-ways) movement. Second, with a maximum payload of 2.2 kg per arm, DE NIRO is limited to relatively light weight tasks, e.g. incapable of lifting a human body. Third, due to limited sensor capabilities in the current design, we constrain DE NIRO to trajectories using forward motion which can result in the robot getting stuck in corners.

DE NIRO can, however, go further. Future work may explore increased awareness, such as through safety improvements with a 360-degree camera rig, the application of a 3D LIDAR (already operational), more robust localization that does not require predefined mapping \cite{bloesch2018codeslam}, human pose estimation, visuospatial skill learning by demonstration \cite{Ahmadzadeh2013IROS}, a more persistent autonomy during navigation without deadlock situations \cite{Kormushev_2016_book_chapter}, and further improvements to point cloud based object detection \cite{gajewski2018adapting} \cite{rusu2008towards}. The work we have accomplished here, nevertheless, shows that DE NIRO's current capabilities can be used to provide reliable, efficient support to tasks requiring frequent, natural interaction with humans.  


\bibliographystyle{IEEEtran}
\bibliography{IEEEabrv,9_bib}

\end{document}